\documentclass[10pt,twocolumn,letterpaper]{article}

\usepackage{cvpr}              

\usepackage[dvipsnames]{xcolor}

\definecolor{cvprblue}{rgb}{0.21,0.49,0.74}
\usepackage[pagebackref=true,breaklinks=true,letterpaper=true,colorlinks,bookmarks=false,citecolor=blue,linkcolor=blue,urlcolor=blue]{hyperref}

\usepackage{times}
\usepackage{epsfig}
\usepackage{graphicx}
\usepackage{amsmath}
\usepackage{amssymb}
\usepackage{booktabs}
\usepackage{multirow}
\usepackage{float}
\usepackage{multicol}
\usepackage{tikz}
\usepackage{comment}
\usepackage{wrapfig}
\usepackage{graphicx}
\usepackage[nopar]{lipsum}
\usepackage{stfloats}
\usepackage{amsmath,amssymb} 
\usepackage{amsfonts}

\usepackage{color}
\usepackage{subcaption}
\usepackage[misc]{ifsym}

\usepackage{orcidlink}
\usepackage[capitalize]{cleveref}
\usepackage{xcolor}
\usepackage{paralist}

\usepackage[capitalize]{cleveref}
\crefname{section}{Sec.}{Secs.}
\Crefname{section}{Section}{Sections}
\Crefname{table}{Table}{Tables}
\crefname{table}{Tab.}{Tabs.}

\def\paperID{145} 
\def\confName{3DV\xspace}
\def\confYear{2024\xspace}

\begin{document}

\title{Physics-based Indirect Illumination for Inverse Rendering}

\author{Youming Deng\\
Cornell University\\
\and
Xueting Li\\
NVIDIA Research\\
\and
Sifei Liu\\
NVIDIA Research\\
\and
Ming-Hsuan Yang\\
UC Merced\\
}
\twocolumn[{%
\renewcommand\twocolumn[1][]{#1}%
\maketitle
 \begin{center}
    \vspace{-5mm}
     \centering
     \includegraphics[width=\textwidth]{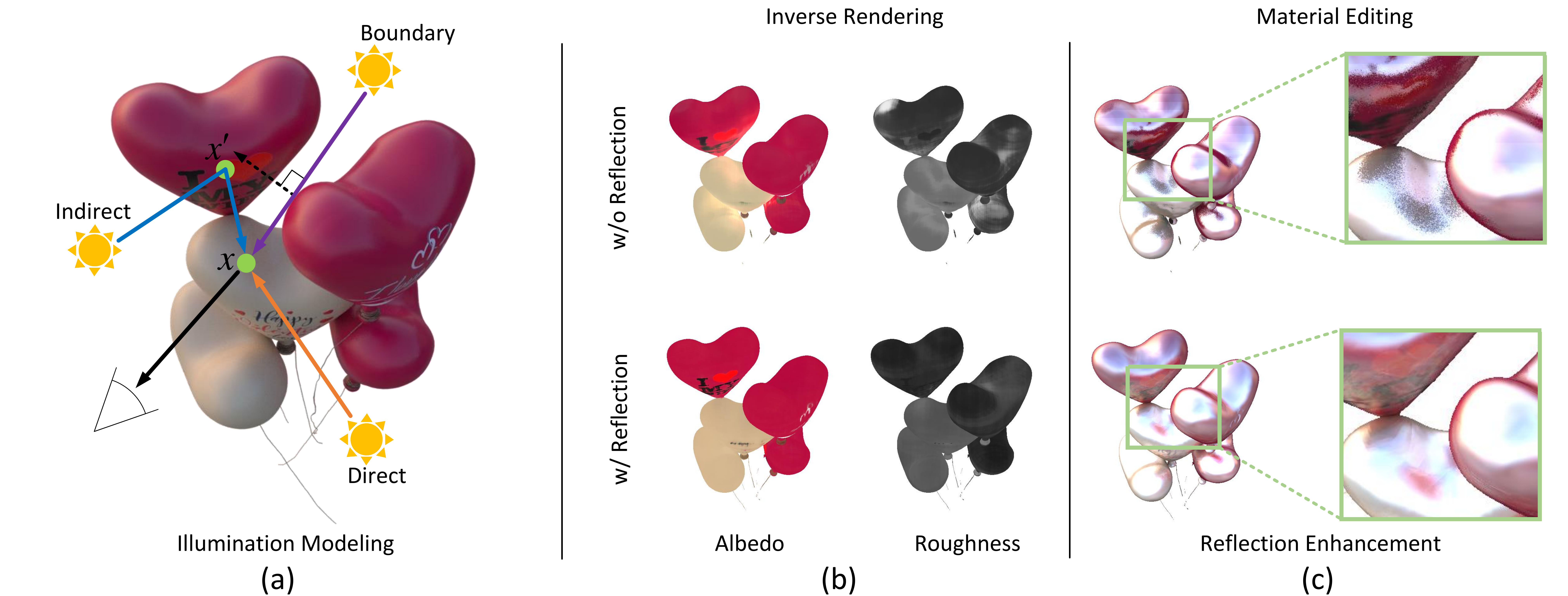}
     \captionof{figure}{(a) Proposed illumination model. (b) Inverse rendering results by our illumination model and a baseline model~\cite{c4} that does not explicitly model reflection. (c) Reflection enhancement. The effectiveness of explicit reflection modeling can be best observed in (c) with edited strong reflection. Our model successfully captures the balloon's reflection, while baseline~\cite{c4} predicts noisy artifacts, as shown in the green box.}
     \label{fig:teaser}
 \end{center}%
}]

\maketitle
\newcommand*{\XT}[1]{\textcolor{orange}{[XT: #1]}}
\newcommand*{\SL}[1]{\textcolor{red}{[SL: #1]}}
\newcommand*{\YM}[1]{\textcolor{green}{[YM: #1]}}
\makeatletter
\newcommand*{\rom}[1]{\expandafter\@slowromancap\romannumeral #1@}
\makeatother

\begin{abstract}
\vspace{-2mm}
    We present a physics-based inverse rendering method that learns the illumination, geometry, and materials of a scene from posed multi-view RGB images.
    To model the illumination of a scene, existing inverse rendering works either completely ignore the indirect illumination or model it by coarse approximations, leading to sub-optimal illumination, geometry, and material prediction of the scene.
    In this work, we propose a physics-based illumination model that first locates surface points through an efficient refined sphere tracing algorithm, then explicitly traces the incoming indirect lights at each surface point based on reflection. Then, we estimate each identified indirect light through an efficient neural network.
    Moreover, we utilize the Leibniz's integral rule to resolve non-differentiability in the proposed illumination model caused by boundary lights inspired by differentiable irradiance in computer graphics.
    As a result, the proposed differentiable illumination model can be learned end-to-end together with geometry and materials estimation.
    As a side product, our physics-based inverse rendering model also facilitates flexible and realistic material editing as well as relighting.
    Extensive experiments on synthetic and real-world datasets demonstrate that the proposed method performs favorably against existing inverse rendering methods on novel view synthesis and inverse rendering.
    The source code and trained models can be found at the project page: \url{https://denghilbert.github.io/pii}.
\end{abstract}
\section{Introduction}
Inverse rendering aims to recover the illumination, geometry, and material simultaneously of a scene from multiview images. 
Illumination plays an important role in inverse rendering by providing crucial lighting information for material and geometry estimation.
However, it is challenging to accurately estimate illumination in a scene due to complex scene geometry and various materials that lead to light occlusion and reflection. 
%
Early inverse rendering works~\cite{c23,c24} adopt a simple illumination model~\cite{c32} and can only be applied to scenes with simple geometry (\eg, scenes that only include a single object).
Some recent approaches~\cite{c36,c26,c4,c27,c37,c10} simulate natural illumination and enable material editing and relighting by utilizing the rendering equation~\cite{c7}.
However, most of these methods~\cite {c26,c37,c10} only model direct illumination and ignore indirect environment lights caused by occlusion and reflection, leading to poor performance on self-occluded scenes.
Even when indirect lights are implicitly modeled as in~\cite{c4,c27,c36,c35}, they are approximated with multi-layer perceptrons (MLPs), ignoring natural occlusion and reflection in the scene. 
One example is shown in the top row of \cref{fig:teaser}b and \cref{fig:teaser}c where the baseline model~\cite{c4} fails to capture reflection and produces noticeable artifacts in both inverse rendering and material editing.

\begin{figure*}[ht]
  \centering
  \includegraphics[width=1\linewidth]{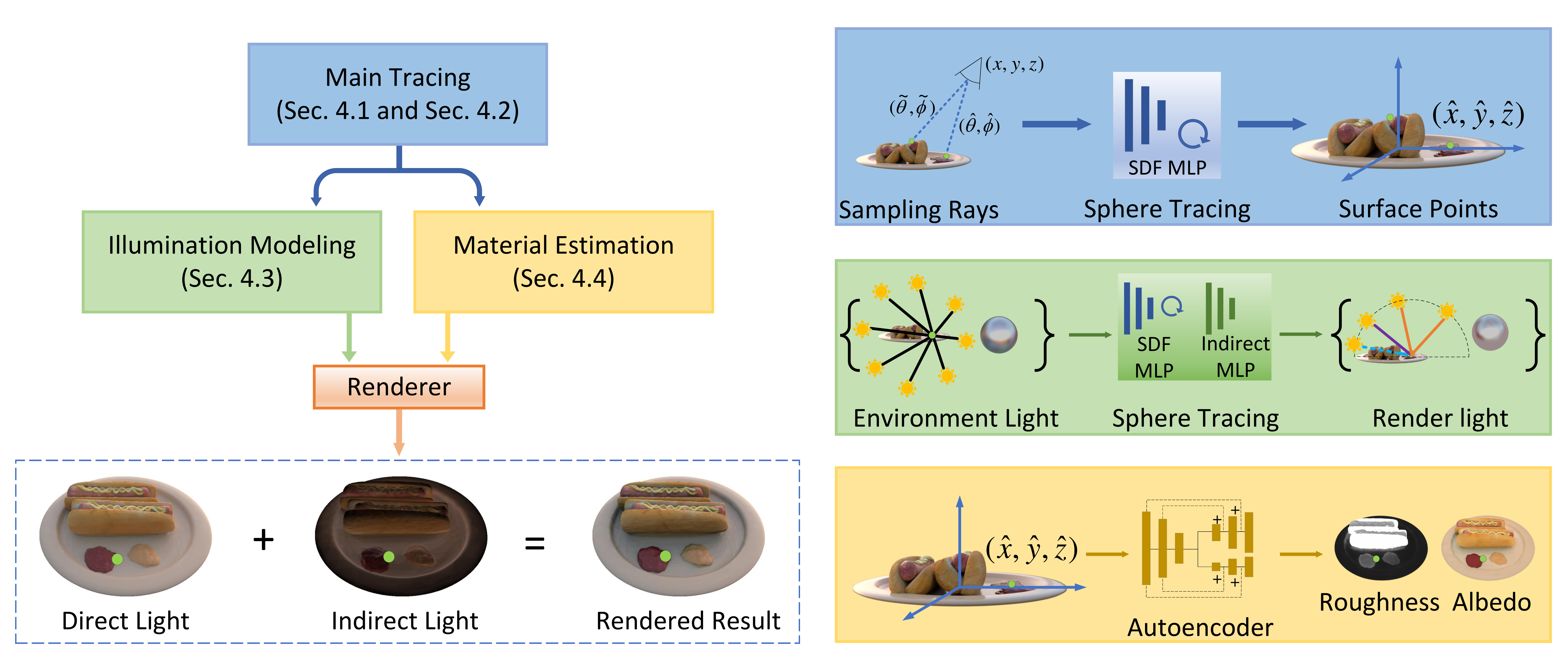}
    \vspace{-7mm}
  \caption{\textbf{Forward Rendering of Our Framework}. Given the camera parameters of a scene view, we sample a set of rays and trace the surface points hit by the rays by the refined sphere tracing module introduced in \cref{sec:tracing refine}. We estimate each traced point's illumination through the illumination module in \cref{sec:indirect} and materials by the material estimation module in \cref{sec:material}. Finally, we feed the estimated materials and illuminations to a differentiable renderer and render a target view image.}
    \vspace{-2mm}
  \label{fig:pipeline}
\end{figure*}

We propose a physics-based inverse rendering framework to accurately model indirect illumination and simultaneously estimate the material and geometry of a scene.
%
For indirect illumination modeling, our method first locates all surface points by a novel and efficient refined sphere tracing algorithm. It then explicitly traces the incoming indirect lights at each surface point based on reflection. Finally, we propose to estimate each traced indirect light by aggregating environment lights using weights predicted by an MLP. However, such a physics-based indirect illumination model is \emph{not} inherently differentiable due to boundary lights~\cite{li2018differentiable,zhang2019differential} -- the environment lights that fall rightly onto the brim of an object (purple ray in \cref{fig:teaser}a). Inspired by differentiable irradiance in graphics~\cite{cermelli2005transport,zhang2019differential}, we derive the gradients of boundary lights theoretically by using the Leibniz's integral rule.
%
%
%
%
%
%
As a result, the proposed illumination module (green block of \cref{fig:pipeline}) can be learned end-to-end together with an SDF prediction MLP (blue block of \cref{fig:pipeline}) and a material estimation autoencoder (yellow block of \cref{fig:pipeline}). 
%
During inference, our method can be readily applied to novel view synthesis, material editing, and relighting, showing favorable performance compared to existing methods on both synthetic and real-world datasets. 

The main contributions of this work are:
\begin{compactitem}
    \item We present a method for high-fidelity geometry, materials, and illumination estimation.
    \item 
    We develop an efficient sphere tracing algorithm for fast surface point locating and an end-to-end differentiable method that models indirect illumination by explicitly tracing and estimating the incoming indirect lights at each surface point.
    \item The proposed method performs favorably against existing methods on inverse rendering, novel-view synthesis, and material editing.
\end{compactitem}
\section{Related Work}
\label{sec:relate}
\noindent \textbf{Implicit Neural Networks.} 
Implicit neural networks encode the geometry of a scene by predicting either occupancy or a signed distance of each surface point. 
Early implicit neural networks~\cite{c20,c21} are learned with ground truth occupancy or signed distances, while more recent models~\cite{c22,c31} are trained with multi-view images by utilizing volume rendering~\cite{c41}. 
However, by entangling geometry with appearance and illumination in neural networks, these methods do not perform well on tasks such as material editing and free-viewpoint relighting. 
In this paper, we disentangle the geometry, material, and illumination by learning an inverse rendering framework. The geometry is captured by an implicit neural network that predicts the signed distance value for each surface point.



\noindent\textbf{Intrinsic Decomposition and Inverse Rendering.} Intrinsic image decomposition~\cite{c32} aims to model an image as the product of reflectance and shading. 
This is a highly ill-posed problem if ground truth annotation is not available. 
As a result, deep intrinsic decomposition models usually resort to synthetic dataset~\cite{bi2018deep,fan2018revisiting,han2018learning,lettry2018darn,NMY:ICCV:2015} or time-lapse sequences~\cite{li2018cgintrinsics}. 
To better model image formulation and enable self-supervised decomposition from images, more recent methods further decompose shading into illumination and geometry, \ie, inverse rendering. 
One line of self-supervised inverse rendering work~\cite{tewari2017mofa,wu2020unsup3d} utilizes the prior of a specific object category (\eg, faces, cars, etc) to constrain illumination prediction. 
To apply inverse rendering on more general scenes, several methods use multi-view images as supervision~\cite{c34,c25}. 
Recently, PhySG~\cite{c10} and NeRFactor~\cite{c42} merge implicit geometry with the rendering equation~\cite{c7} to decompose material properties and illuminations from visual appearance.
PhySG~\cite{c10} resolves inverse rendering by modeling illumination as mixtures of spherical Gaussians~\cite{c8}, combined with an implicit function for geometry estimation.
However, PhySG does not explicitly model indirect illumination, which plays an important role in capturing reflection in a scene. 
To resolve this issue, several approaches~\cite{c4,c27,c35,c36,c42} approximate the indirect illumination and visibility in the scene through MLPs. Other methods use more physical-based modeling~\cite{wu2023nefii,liu2023nero}.
Nonetheless, without explicitly tracing and estimating indirect environment lights, these methods do not accurately capture the reflections in the scene and produce artifacts in both inverse rendering and material editing.
In contrast, we propose a self-supervised inverse rendering method that explicitly traces and estimates the indirect lights at each surface point, showing promising results in both inverse rendering and material editing.

\section{Preliminaries}
\label{sec:pre}

\subsection{Rendering Equation}
\label{sec:rendering}
Before describing our method, we briefly review the Bidirectional Reflectance Distribution Function (BRDF) and the rendering equation that computes outgoing radiance at each surface point in the scene. 

Given a surface point $\boldsymbol{x}$ and the outgoing light direction $\boldsymbol{\omega}_o$, the rendering equation~\cite{c7} computes the outgoing light $L_o(\boldsymbol{\omega}_o;\boldsymbol{x})$ as the integral over all environment lights $L_i(\boldsymbol{\omega}_i)$ from the upper hemisphere $\Omega$:
\begin{equation}
\label{eq:render equ}
    L_o(\boldsymbol{\omega}_o; \boldsymbol{x})=\int_\Omega L_i(\boldsymbol{\omega}_i)f_{r}(\boldsymbol{\omega}_o,\boldsymbol{\omega}_i;\boldsymbol{x})(\boldsymbol{\omega}_i\cdot \boldsymbol{n}){\rm d}\boldsymbol{\omega}_i,
\end{equation}where $f_{r}(\boldsymbol{\omega}_o,\boldsymbol{\omega}_i;\boldsymbol{x})$ is the simplified Disney BRDF~\cite{c9} that models the reflectance property of the surface point $\boldsymbol{x}$ and $\boldsymbol{n}$ is the normal at this point. The specularity term in $f_{r}(\boldsymbol{\omega}_o,\boldsymbol{\omega}_i;\boldsymbol{x})$ is modeled with the microfacet model~\cite{c28,c29}. More details can be found in the supplementary. 

\subsection{Spherical Gaussians Function}
\label{sec:sg}
We model the environment light $L_i(\boldsymbol{\omega}_i)$,  BRDF $f_{r}(\boldsymbol{\omega}_o,\boldsymbol{\omega}_i;\boldsymbol{x})$, and multiplication term $\boldsymbol{\omega}_i\cdot \boldsymbol{n}$ in \cref{eq:render equ} by the Spherical Gaussian (SG) Distribution~\cite{c8}.
A Spherical Gaussian is defined as:
\begin{equation}
    G(\boldsymbol{\omega}_i;\boldsymbol{\xi},\lambda,\boldsymbol{\mu})=\boldsymbol{\mu}e^{\lambda(\boldsymbol{\omega}_i\cdot\boldsymbol{\xi}-1)},
\end{equation}where $\boldsymbol{\omega}_i\in \mathbb{S}^2$ is the input of the SG function, $\boldsymbol{\xi}\in \mathbb{S}^2$ is the lobe axis, $\lambda\in\mathbb{R}_+$ is the lobe sharpness, and $\boldsymbol{\mu}\in\mathbb{R}^3$ is the lobe amplitude.

When modeling the environment lights using the SG function, we assume the lights are uniformly cast from 128 infinitely faraway directions. 
We analyze the number of environment lights in \cref{sec:ablation}. 
For computational efficiency, we further represent the BRDF $f_{r}$, and the multiplication term $\boldsymbol{\omega}_i\cdot \boldsymbol{n}$ in \cref{eq:render equ} with SG as in~\cite{c10}. 
More details can be found in the supplementary material.

\section{Proposed Method}
\label{sec:method}

Given multiple images of a scene, we learn a self-supervised inverse rendering framework that estimates the geometry, materials, and illumination of the scene. 
During inference, our model is readily available for novel view synthesis, relighting, and material editing. 
We show the overview of our pipeline in \cref{fig:pipeline} and introduce details for geometry estimation (\cref{sec:sdf mlp}), surface point localization (\cref{sec:tracing refine}), illumination modeling (\cref{sec:indirect}) and materials prediction (\cref{sec:material}) below.


\subsection{Geometry Estimation}
\label{sec:sdf mlp}
We encode the geometry of a given scene by learning an MLP $F_{SDF}$ that takes a query point $\boldsymbol{p}=(x,y,z)$ as input and predicts its SDF value (\ie, the closest distance between $\boldsymbol{p}$ and the object surface).
We first initialize this geometry MLP using multi-view images as in~\cite{c26,c4,c27}. 
We then fine-tune the MLP with the illumination and material estimation modules in the following sections.

\subsection{Sphere Tracing with Refinement}
\label{sec:tracing refine}

We discuss locating all surface points using the learned geometry MLP discussed in~\cref{sec:sdf mlp} and a novel refined sphere tracing module.

\begin{figure}[t]
  \centering
  \includegraphics[width=1\linewidth]{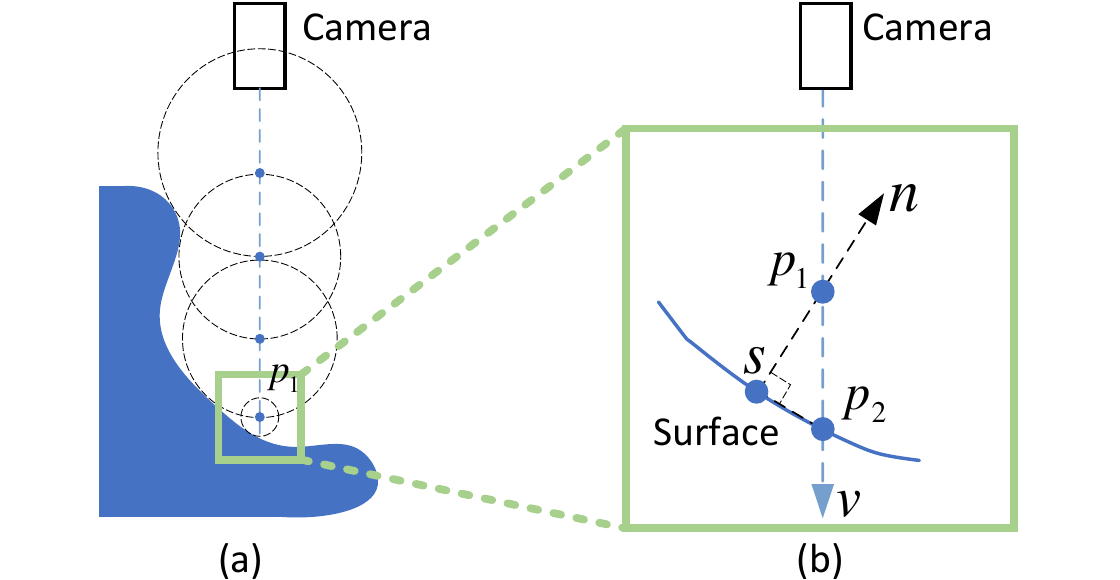}
  \caption{\textbf{Sphere Tracing with Refinement}. The classic sphere tracing algorithm stops at $\boldsymbol{p}_1$ when the SDF value of this point is smaller than the threshold. We propose a refinement process that directly marches from $\boldsymbol{p}_1$ to point $\boldsymbol{p}_2$ that is closer to the surface.}
  \label{fig:tracing}
  \vspace{-2mm}
\end{figure}

Given camera parameters, classical sphere tracing algorithms~\cite{c5,c17} trace all surface points by marching along sampled rays iteratively until reaching a point with an SDF value smaller than a given threshold $\tau$.
We demonstrate this tracing process in \cref{fig:tracing}(a) and show the traced surface point $\boldsymbol{p}_1$ as well as the ``true'' surface point $\boldsymbol{p}_2$ in \cref{fig:tracing}(b). 
As shown in \cref{fig:tracing}, a larger threshold $\tau$ leads to a more inaccurate surface point location (\ie, $\boldsymbol{p}_1$ is farther from the object surface), while a smaller threshold requires more tracing iterations. To efficiently and accurately locate a surface point, we propose a refinement process that directly steps from $\boldsymbol{p}_1$ to $\boldsymbol{p}_2$. Our main observation is that as $\boldsymbol{p}_1$ moves closer to $\boldsymbol{p}_2$, the surface curve $\boldsymbol{s}\boldsymbol{p}_2$ approximates to a straight line. 
Thus we can find the length of $\overline{\boldsymbol{p}_1\boldsymbol{p}_2}$ by:
\begin{equation}
    \overline{\boldsymbol{p}_1\boldsymbol{p}_2}=\frac{-F_{SDF}(\boldsymbol{p}_1)}{\boldsymbol{v}\cdot \boldsymbol{n}}, \ \  \boldsymbol{n}=\frac{\nabla_{\boldsymbol{p}}F_{SDF}(\boldsymbol{p})}{||\nabla_{\boldsymbol{p}}F_{SDF}(\boldsymbol{p})||_2},
\end{equation}
where $\boldsymbol{v}$ is the normalized view direction, and the normal $\boldsymbol{n}$ at point $\boldsymbol{s}$ can be derived from the learned geometry MLP $F_{SDF}$ discussed in \cref{sec:sdf mlp}.

Compared to classic sphere tracing methods~\cite{c5,c17}, the proposed refinement step uses a relatively large threshold and requires fewer tracing iterations and less time. We demonstrate its effectiveness in Sec.~\ref{sec:ablation}.

\subsection{Illumination Modeling}
\label{sec:indirect}

Illumination plays a critical role in recovering the geometry and materials of a scene from a set of RGB images. 
However, it is challenging to accurately model the environment lights due to indirect reflection. 
Existing approaches either ignore the indirect lights~\cite{c26,c37,c10} or implicitly approximate them~\cite{c4,c27,c36} via neural networks.
Instead, we explicitly trace and estimate the incoming indirect lights at each surface point based on occlusion and reflection.
In the following, we only focus on indirect lights and omit direct lights that are straightforward to model by~\cref{eq:render equ}.
%
%

\vspace{-1.5mm}
\subsubsection{Indirect Lights Identification} 
\label{sec: classification}
For each environment light, we first classify it as a direct or indirect light by a simplified ray tracing process that explicitly models reflection.
%

Given a surface point $\boldsymbol{x}$, we reversely trace each environment light starting from $\boldsymbol{x}$ to the light sources. 
Rays not obstructed by an object are classified as direct lights (orange rays in \cref{fig:teaser}(a)), and obstructed rays are denoted as indirect lights (blue rays in \cref{fig:teaser}(a)).
%
We note that here we do not trace the full path of indirect light as classic ray tracing methods~\cite{c40}. 
As long as a ray starting from the surface point hits an object while reaching the light sources, we classify the environment light as an indirect light and model it using the method discussed in \cref{sec:indirect light}. 
Although this efficient tracing process is simpler compared to methods that trace the complete path, we demonstrate its effectiveness with extensive experiments in \cref{sec:experiment}.


\begin{figure}[t]
    \centering
    \includegraphics[width=1\linewidth]{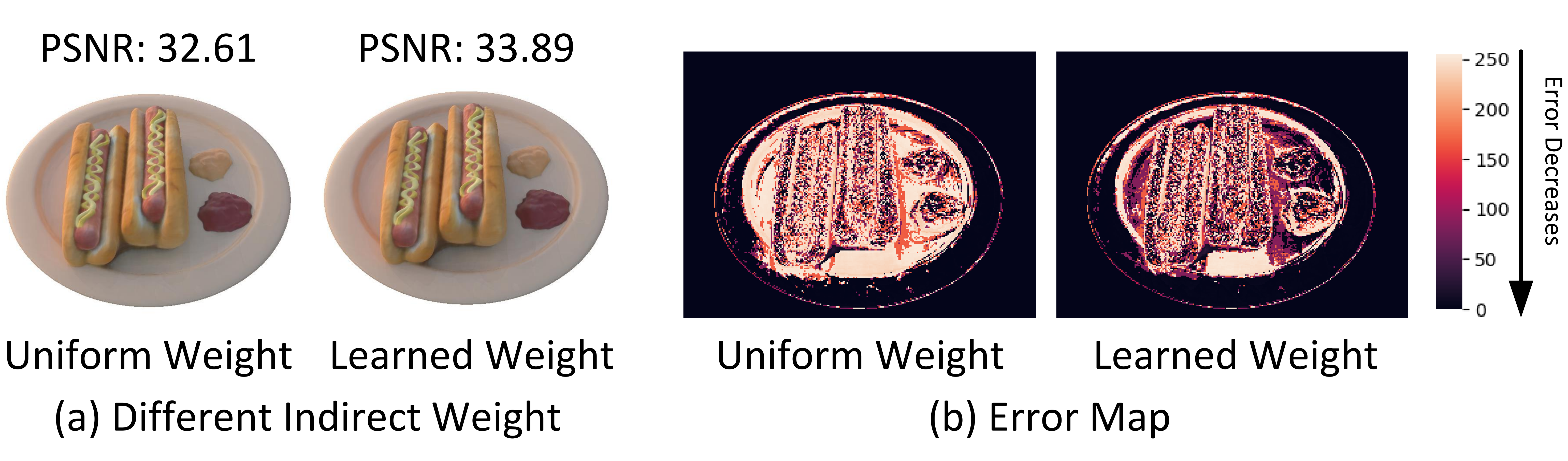}
    \vspace{-7mm}
    \caption{\textbf{Environment Weight}. We visualize rendering results and their error maps with uniform and learned weights.}
    \label{fig:visual}
    \vspace{-2mm}
\end{figure}
\subsubsection{Indirect Lights Estimation}
\label{sec:indirect light}

Given each identified indirect light at the surface point $\boldsymbol{x}$ coming from a reflected point $\boldsymbol{x}'$ of \cref{fig:teaser}(a), we learn an MLP, $F_{ind}$, that explicitly formulates the indirect light by aggregating environment lights. 
Specifically, for a target indirect light $L_r$, the MLP takes the coordinates and normal of the reflected point $\boldsymbol{x}'$ as inputs and predicts how much each environment light $L^i$ contributes to $L_r$ represented as a scalar weight $w_{r}^i \in \mathbb{R}$. Thus, the lobe sharpness $\lambda_r$ and amplitude $\boldsymbol{\mu}_r$ of $L_r$ can be computed as:
\begin{gather}
    \lambda_r = \sum_{i=1}^{K} w_{r}^i\cdot\lambda^i, \boldsymbol{\mu}_r=\sum_{i=1}^{K} w_{r}^i\cdot\boldsymbol{\mu}^i,\\
    \boldsymbol{w_{r}}=F_{ind}(\gamma(\boldsymbol{x}'),\gamma(\boldsymbol{n}')), \label{eq:weight}
\end{gather}
where $\boldsymbol{n}'$ is the normal at $\boldsymbol{x}'$ with positional embedding $\gamma$~\cite{c12}, and $K$ is the total environment light number (\eg, 128 in this work). We carry out further ablation study in \cref{sec:ablation} to verify the effectiveness of this integral weight design on modeling indirect illumination. 

Overall, the proposed indirect light model has two advantages compared to existing inverse rendering methods~\cite{c4,c27,c35,c36}. 
First, reflection in the scene is explicitly modeled through the explicit tracing and classification process discussed in \cref{sec: classification}. 
%
Second, instead of simply approximating all indirect lights at the surface point (\ie $x$ in~\cref{fig:teaser}) by an MLP~\cite{c4}, we trace to find reflected points (\ie $x'$ in~\cref{fig:teaser}) and estimate each identified indirect light through learnable integral weights (\ie $w_r$ in ~\cref{eq:weight}), leading to an illumination model with a higher capacity to handle reflection.
 \cref{fig:visual}, \cref{fig:qualitative synthetic}, and \cref{fig:material edit} empirically demonstrate the superiority of our indirect light model.
%

\begin{figure}
    \centering
    \includegraphics[width=1\linewidth]{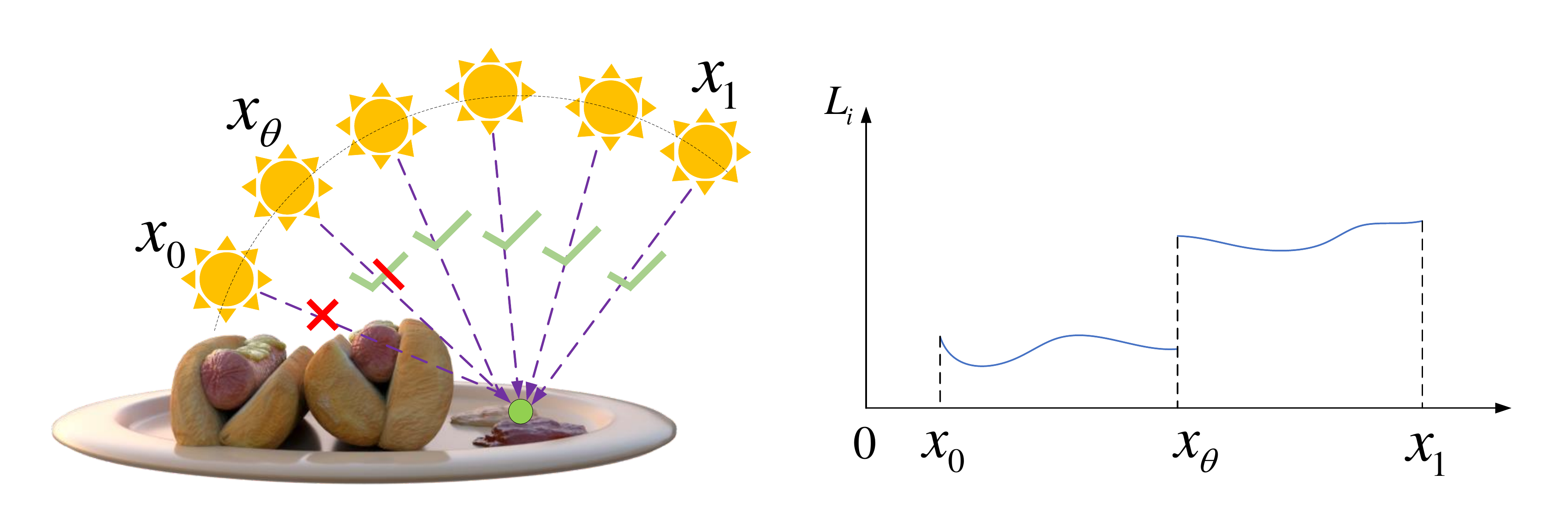}
    \vspace{-7mm}
    \caption{\textbf{Demonstration of Non-differentiability at Boundary Lights.} The boundary light $x_{\theta}$ changes abruptly from ``obstructed'' to ``visible'' at the boundary of the object, introducing a point of discontinuity and making \cref{eq:render equ} non-differentiable. This non-differentiability is also demonstrated by the figure on the right showing light intensity change w.r.t. the incoming light angle $x_i$.}
    \label{fig:indiff}
    \vspace{-2mm}
\end{figure}

\subsubsection{Boundary Lights Differentiation}
\label{sec:tangent light}
Ideally, the identified direct lights in~\cref{sec: classification} and estimated indirect lights in~\cref{sec:indirect light} can be aggregated by ~\cref{eq:render equ} and learned through backpropagation.
%
However, the boundary lights that fall right onto the brim of an object make \cref{eq:render equ} non-differentiable since they change abruptly at the object boundary from ``obstructed'' to ``visible'', as shown in \cref{fig:indiff}. Naively ignoring the boundary lights does not consider errors from boundary lights and leads to inaccurate illumination estimation (see the last row of \cref{tab:Light Estimation}).
%
%
%
%
Inspired by recent advances of differential irradiance in computer graphics~\cite{cermelli2005transport,zhang2019differential}, we use the Leibniz's integral rule and compute the gradient of the outgoing radiance \wrt all learnable parameters $\boldsymbol{\theta}$ as:
\begin{equation}
\label{eq:silhouette}
\begin{aligned}
        &\frac{\partial L_o(\boldsymbol{\omega}_o; \boldsymbol{x})}{\partial \boldsymbol{\theta}}=\int_{\Omega'}\frac{\partial L_i(\boldsymbol{\omega}_i)f_{r}(\boldsymbol{\omega}_o,\boldsymbol{\omega}_i;\boldsymbol{x})}{\partial\boldsymbol{\theta}}(\boldsymbol{\omega}_i\cdot \boldsymbol{n}){\rm d}\boldsymbol{\omega}_i\\&+\int_{\partial\Omega}(\frac{\partial\boldsymbol{\omega}_i}{\partial\boldsymbol{\theta}}\cdot \boldsymbol{n}(\boldsymbol{\omega}_i)) \Delta L_i(\boldsymbol{\omega}_i)f_{r}(\boldsymbol{\omega}_o,\boldsymbol{\omega}_i;\boldsymbol{x})(\boldsymbol{\omega}_i\cdot \boldsymbol{n}){\rm d}\boldsymbol{\omega}_i,
\end{aligned}\end{equation}
where $\partial \Omega$ represents the edge ($x_{\theta}$ in \cref{fig:indiff}) at which the light changes from ``obstructed" to ``visible", $\Delta L_i(\boldsymbol{\omega}_i)=L_i(\boldsymbol{\omega}_i+\boldsymbol{\epsilon})-L_i(\boldsymbol{\omega}_i-\boldsymbol{\epsilon})$ is the light difference of directions on $\partial\Omega$, and $\Omega'=\Omega-\partial\Omega$. The supplementary material presents the detailed mathematical deduction of \cref{eq:silhouette}. 

To identify the boundary lights, we simply classify environment lights that are ``almost'' perpendicular (\eg, a 2-degree deviation) to the surface normal.


\subsection{Material Estimation}
\label{sec:material}
Similar to~\cite{c4}, we predict its roughness and albedo using a material autoencoder~\cite{c18} at each point, as shown in the yellow block of \cref{fig:pipeline}.
Specifically, we first map the positional encoding~\cite{c12} of the surface point coordinates into a latent vector through the material encoder. 
Then we decode the latent vector into roughness and albedo using two separate decoders. 
Given the estimated materials and illumination discussed in~\cref{sec:indirect}, we then render the scene from any given view direction with the rendering equation~\cite{c7}.

\subsection{Training}
\label{sec:training}
We pre-train the geometry MLP $F_{SDF}$ discussed in \cref{sec:sdf mlp} following the MII~\cite{c4}. 
This provides decent geometry information for the material and illumination module training.
Next, we jointly train the geometry MLP, the material, and the illumination module by minimizing the L1 reconstruction loss $\ell_{rec}$.
%
We further encourage the material autoencoder to produce smooth roughness and albedo by applying a latent sparsity constraint with KL divergence $\ell_{KL}$~\cite{c19}, and a smoothness loss $\ell_{smooth}$~\cite{c4}.
The overall objective is:
\begin{gather}
\ell  =\lambda_{rec}\ell_{rec}+\lambda_{KL}\ell_{KL}+\lambda_{smooth}\ell_{smooth},\\
    \ell_{KL} =\sum_{i=1}^N KL(\rho||z_i),\\ \ell_{smooth} =||D(\boldsymbol{\boldsymbol{z}})-D(\boldsymbol{\boldsymbol{z}}+\boldsymbol{\epsilon})||,
\end{gather} 
where $N$ is the dimension of latent vector $\boldsymbol{z}$, and $D$ represents the decoders in the material autoencoder. 
We set $\rho$ to be 0.05 and $\boldsymbol{\epsilon}$ to be 0.02. 
The weights are set to be $\lambda_{rec}=1.0$, $\lambda_{KL}=0.01$, $\lambda_{smooth}=0.1$.

To summarize, given multi-view images of a scene, our framework models its: (1) \textit{illumination} via a set of learnable environment lights $L^i$ represented as SGs $\{\boldsymbol{\xi}_i,\lambda_i,\boldsymbol{\mu}_i\}^{128}_{i=1}$, (2) \textit{geometry} via an MLP $F_{SDF}$, (3) \textit{material} property (\ie, albedo $\boldsymbol{a}$ and roughness $R$ for each surface point) via a material autoencoder.

\begin{table*}[ht]
    \begin{subtable}[h]{0.99\textwidth}
        \centering
    \scalebox{0.83}{
     \begin{tabular}{ccccccccccccc}
    \toprule
    \multirow{2}[1]{*}{\#Train/Test (HxW)} & \multicolumn{3}{c}{Synthetic Chair} & \multicolumn{3}{c}{Synthetic Jugs} & \multicolumn{3}{c}{Synthetic Air Balloons} & \multicolumn{3}{c}{Synthetic Hotdog} \\
          & \multicolumn{3}{c}{200/200 (800x800)} & \multicolumn{3}{c}{100/200 (800x800)} & \multicolumn{3}{c}{100/200 (800x800)} & \multicolumn{3}{c}{100/200 (800x800)} \\
    Metrics & $\downarrow$LPIPS & $\uparrow$SSIM  & $\uparrow$PSNR & $\downarrow$LPIPS & $\uparrow$SSIM  & $\uparrow$PSNR & $\downarrow$LPIPS & $\uparrow$SSIM  & $\uparrow$PSNR &  $\downarrow$LPIPS & $\uparrow$SSIM  & $\uparrow$PSNR \\
    \midrule
    PhySG*~\cite{c10} & 0.0406 & 0.9497 & 32.08 & 0.0515 & 0.9742 & 33.59  & 0.0439 & \textbf{0.9618} & 31.79 & 0.0317 & 0.9564 & 31.95 \\
    MII~\cite{c4}   & 0.0147 & 0.9482 & 35.42 & 0.0193 & 0.9714 & 37.47 & 0.0468 & 0.9498 & 31.03 & 0.0216 & 0.9531 & 32.69 \\
    \midrule
    Ours (Learned) & \textbf{0.0125} & \textbf{0.9534} & \textbf{36.04} & \textbf{0.0148} & \textbf{0.9788} & \textbf{38.76} &  \textbf{0.0364} &  0.9613 & \textbf{32.72}   &  \textbf{0.0192}     &    \textbf{0.9589}   &  \textbf{33.76} \\
    Ours (Uniform)  & 0.0173 & 0.9526 & 33.62 & 0.0301 & 0.9728 & 33.79 &  0.0487 &  0.9566 & 30.13   &  0.0221     &    0.9564  &  32.74 \\
    \bottomrule
    \end{tabular}%
    }
    
    \caption{\textbf{Novel View Synthesis}.}
    \label{tab:novel view}%
    \end{subtable}
    \begin{subtable}[h]{0.99\textwidth}
        \centering
    \scalebox{0.82}{
     \begin{tabular}{ccccccccccccc}
    \toprule
    \multirow{2}[1]{*}{\#Train/Test (HxW)} & \multicolumn{3}{c}{Synthetic Chair} & \multicolumn{3}{c}{Synthetic Jugs} & \multicolumn{3}{c}{Synthetic Air Balloons} & \multicolumn{3}{c}{Synthetic Hotdog} \\
          & \multicolumn{3}{c}{200/200 (800x800)} & \multicolumn{3}{c}{100/200 (800x800)} & \multicolumn{3}{c}{100/200 (800x800)} & \multicolumn{3}{c}{100/200 (800x800)} \\
    Metrics & $\downarrow$LPIPS & $\uparrow$SSIM  & $\uparrow$PSNR & $\downarrow$LPIPS & $\uparrow$SSIM  & $\uparrow$PSNR & $\downarrow$LPIPS & $\uparrow$SSIM  & $\uparrow$PSNR &  $\downarrow$LPIPS & $\uparrow$SSIM  & $\uparrow$PSNR \\
    \midrule
    PhySG*~\cite{c10} & 0.1004 & \textbf{0.9159} & 24.58 & \textbf{0.2119} & 0.9415 & \textbf{27.38} & 0.1395 & 0.9405 & 24.51 & 0.1517 & 0.9229 & 19.51 \\
    MII~\cite{c4}   & 0.1158 & 0.9063 & 24.96 & 0.2302 & 0.9216 & 26.719 & 0.1636 & 0.9218 & 22.53  & \textbf{0.0770} & 0.9401 & \textbf{23.64} \\
    \midrule
    Ours  & \textbf{0.0881} & 0.9083 & \textbf{25.07} & 0.2138 & \textbf{0.9477} & 26.80 & \textbf{0.1147} & \textbf{0.9513} & \textbf{26.43} & 0.0887 & \textbf{0.9415} & 22.77 \\
    Ours (w/o boundary) & 0.0997 & 0.9131 & 25.01 & 0.2195 & 0.9436 & 26.83 & 0.1173 & 0.9494 & 26.22 & 0.1008 & 0.9345 & 22.72 \\
    \bottomrule
    \end{tabular}%
    }
    \caption{\textbf{Albedo Estimation}.}
    \label{tab:albedo}%
    \end{subtable}
    \begin{subtable}[h]{0.63\textwidth}
        \centering
        \scalebox{0.73}{
            \begin{tabular}{ccccccccc}
        \toprule
        scenes & \multicolumn{2}{c}{Synthetic Chair} & \multicolumn{2}{c}{Synthetic Jugs} & \multicolumn{2}{c}{Synthetic Air Balloons} & \multicolumn{2}{c}{Synthetic Hotdog} \\
        \cmidrule{2-9}    Metrics & $\downarrow$MSE   & $\downarrow$ARE   & $\downarrow$MSE   & $\downarrow$ARE   & $\downarrow$MSE   & $\downarrow$ARE   & $\downarrow$MSE   & $\downarrow$ARE \\
        \midrule
        PhySG*~\cite{c10} & 0.0921 & 1.1771 & 0.1200  & 1.2974 & 0.1832 & 3.0347 & 0.2861 & 24.6206 \\
        MII~\cite{c4}   & \textbf{0.0306} & 1.1861 & 0.0617 & 1.1789 & 0.0242 & 0.8936 & \textbf{0.0776} & 4.7174 \\
        \midrule
        Ours & 0.0421 & \textbf{1.0829} & \textbf{0.0484} & \textbf{0.9123} & \textbf{0.0132}& \textbf{0.6450} & 0.0819 & \textbf{4.5096} \\
        \bottomrule
        \end{tabular}%
          }
        
        \caption{\textbf{Roughness Estimation}.}
        \label{tab:Roughness Estimation}
        \end{subtable}
    \begin{subtable}[h]{0.36\textwidth}
        \centering
        \scalebox{0.63}{
   \begin{tabular}{ccccc}
\toprule
Scenes & Chair & Jugs & Air Balloons & Hotdog \\
\cmidrule{2-5}    Metric & \multicolumn{4}{c}{$\downarrow$Light Estimation MSE} \\
\midrule
PhySG*~\cite{c10} & 0.0557 & 0.1067 & 0.0154 & 0.0493 \\
MII~\cite{c4}   & 0.0562 & 0.1371 & 0.0186 & 0.0256 \\
\midrule
Ours & \textbf{0.0232} & \textbf{0.0555} & \textbf{0.0120} & \textbf{0.0144} \\
Ours (w/o boundary) & 0.0386 & 0.0699 & 0.0213 & 0.0427 \\
\bottomrule
\end{tabular}
  }

\caption{\textbf{Light Estimation}.}
\label{tab:Light Estimation}
\end{subtable}
    \vspace{-2mm}
    \caption{\textbf{Quantitative Results}. We compare our method with PhySG*~\cite{c10} (slightly modified, see \cref{sec:comparison}) and MII~\cite{c4} on four tasks: novel view synthesis, albedo, and roughness estimation, as well as environment map estimation. HDR images used in synthetic data are tone mapped with $I_{new}=I_{old}^{1/2.2}$ and clipped to $\left[0,1\right]$.}
    \label{tab:table1}
    \vspace{-3mm}
\end{table*}%
\section{Experimental Results}
\label{sec:experiment}

We evaluate our method on the synthetic scenes in~\cite{c4}. 
All the scenes are rendered with natural environment lights and provided with foreground masks. 
For quantitative evaluation, the dataset provides 200 testing images along with the ground truth albedo and roughness. 
The images have a resolution of 800$\times$800.

\subsection{Comparisons with State-of-the-art Methods}
\label{sec:comparison}
\noindent\textbf{Baseline methods.} We compare with two state-of-the-art inverse rendering methods: PhySG~\cite{c10} and MII~\cite{c4}. 
Both methods predict illumination, materials, and geometry from multi-view images. 
Following MII~\cite{c4}, we replace the global roughness module in PhySG~\cite{c10} with our spatial-varying material net since the original one cannot estimate spatially-varying surface roughness.

\noindent\textbf{Metrics.} We conduct extensive experiments to compare the proposed method and baselines on novel view synthesis, materials estimation, and environment light prediction. 
Specifically, for novel view synthesis and albedo prediction, we report the PSNR, SSIM, and LPIPS~\cite{c1} between the ground truth and predictions.
For roughness estimation, we adopt the mean square error (MSE) and the absolute relative error~\cite{c3} as evaluation metrics. 
While for illumination estimation, we compare the MSE between the ground truth and predicted environment maps as in~\cite{c2}.

\noindent\textbf{Quantitative Results}. We report quantitative evaluation on novel view synthesis, materials, and illumination evaluation in \cref{tab:table1}. 
We use ground truth images, albedo, roughness, and environment maps to compute the abovementioned metrics. 
As shown in \cref{tab:novel view} and \cref{tab:Light Estimation}, our method consistently outperforms the baseline models on all testing scenes on novel view synthesis and illumination estimation. 
For material estimation, our method performs favorably against baselines on all testing scenes and most metrics quantitatively, as demonstrated in \cref{tab:albedo,tab:Roughness Estimation}. The supplementary material will show the other two air balloons and chairs scenes.

\begin{figure*}[ht]
  \centering
  \scalebox{0.8}{
    \includegraphics[width=1.24\linewidth]{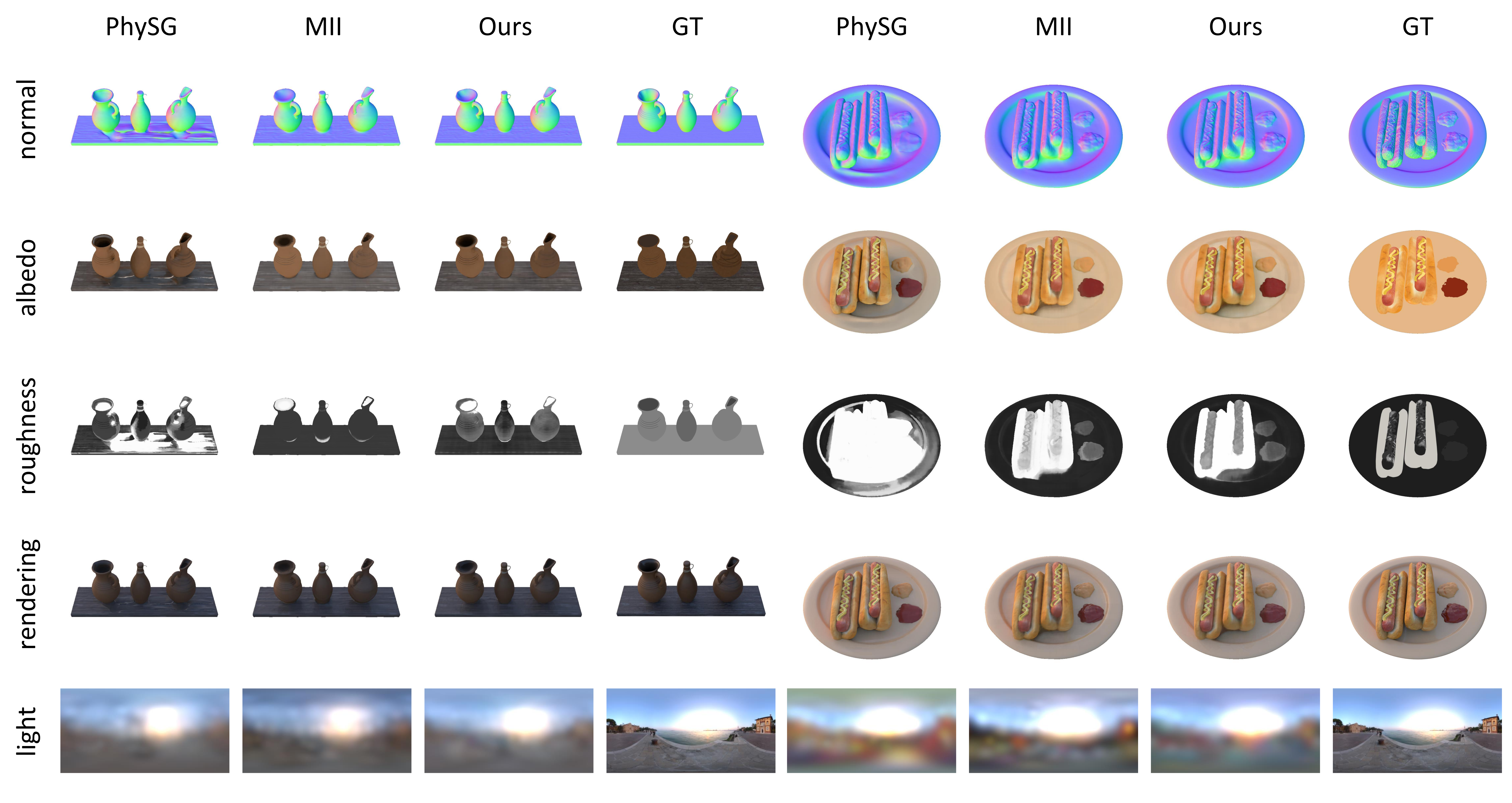}
  }
  \vspace{-5mm}
  \caption{\textbf{Qualitative Comparisons on Synthetic Dataset}. We show qualitative comparisons of PhySG~\cite{c10}, MII~\cite{c4}, and our method on novel view synthesis, albedo, roughness, and environment map estimation. More scenes can be found in the supplementary.}
  \label{fig:qualitative synthetic}
\end{figure*}

\noindent\textbf{Qualitative Results}. 
As shown in \cref{fig:qualitative synthetic}, our model predicts plausible roughness and material while PhySG~\cite{c10} suffers from disentangling illumination and materials where the lights are obstructed (\eg, the occluded jug bottoms and the white balloon in the middle). Furthermore, compared to MII~\cite{c4}, our method estimates more consistent roughness and realistic albedo by explicitly tracing and estimating indirect lights based on occlusion and reflectance.

\begin{figure*}[ht]
    \centering
  \includegraphics[width=1\linewidth]{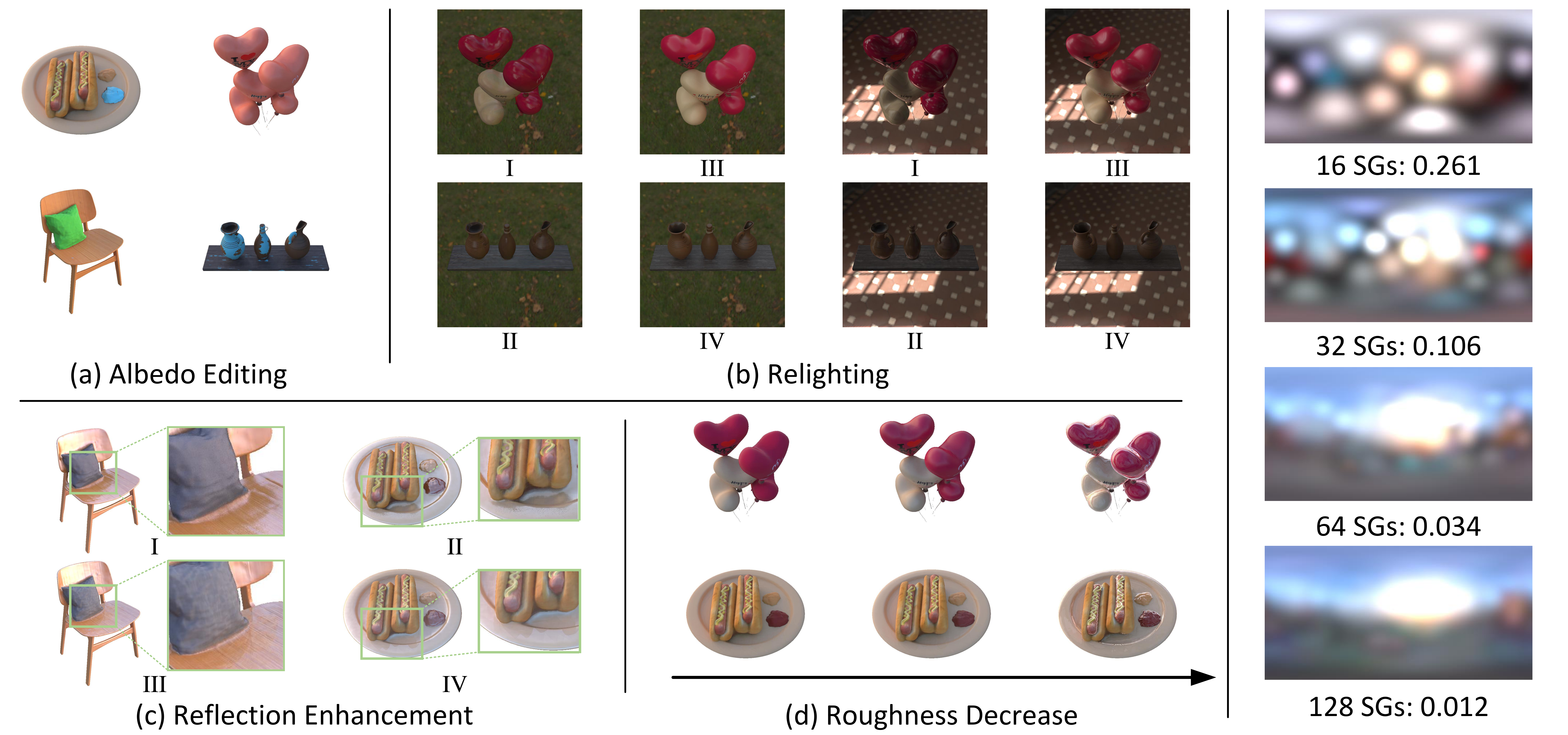}
  \caption{\textbf{Material Editing and Relighting}. \textit{Left:} (a) Base color editing results by changing the albedo. (b) Relighting results (\rom{1}, \rom{2}) by changing the illumination SGs and GT (\rom{3}, \rom{4}) rendered with Blender~\cite{c2}. (c) Enhancement of reflection and comparison between MII~\cite{c4} (\rom{1}, \rom{2}) and our model (\rom{3}, \rom{4}). (d) The gradual decrease of the surface roughness and the objects' surface change from matte to smooth. \textit{Right:} We experiment with different SG numbers for illumination to find a balance between a high-fidelity model and memory efficiency.}
  \label{fig:material edit}
\end{figure*}

\noindent\textbf{Indirect Reflection}. We show that our model effectively captures the indirect reflection in a scene by changing the Fresnel coefficient~\cite{c9} in the Fresnel term\footnote{Fresnel term models how much light leaves the object surface with a given arbitrary direction of incidence.}. 
To this end, we increase the coefficient to enhance surface reflection, as shown in \cref{fig:teaser}c and \cref{fig:material edit}c. 
Surprisingly, our model can capture the indirect reflection in every scene. 
We observe the reflected heart-shaped pattern in the balloon scene in \cref{fig:teaser}c, the shadows of the back of the chair and pillow accurately casting onto the seat, and the precise edge shape of the hotdog projecting on the plate (green boxes in \cref{fig:material edit}c). 
In contrast, the same editing applied to the MII~\cite{c4} generates noisy black points and noticeable artifacts for indirect reflection. 
This empirically demonstrates that the proposed method successfully mimics the reflectance in the physical world, which is rarely achieved by existing models~\cite{c4,c27,c36}.

\noindent\textbf{Relighting and Editing}. Thanks to the rendering equation, our model facilitates flexible editing for both material and illumination during inference.
We demonstrate how to edit the learned albedo in \cref{fig:material edit}a, the environment light in \cref{fig:material edit}b, and the roughness in \cref{fig:material edit}d. Our framework can get reasonable rendering results with different base colors or environment illumination.
More results and videos are available in the supplementary material.

\begin{figure}[ht]
  \centering
  \includegraphics[width=1\linewidth]{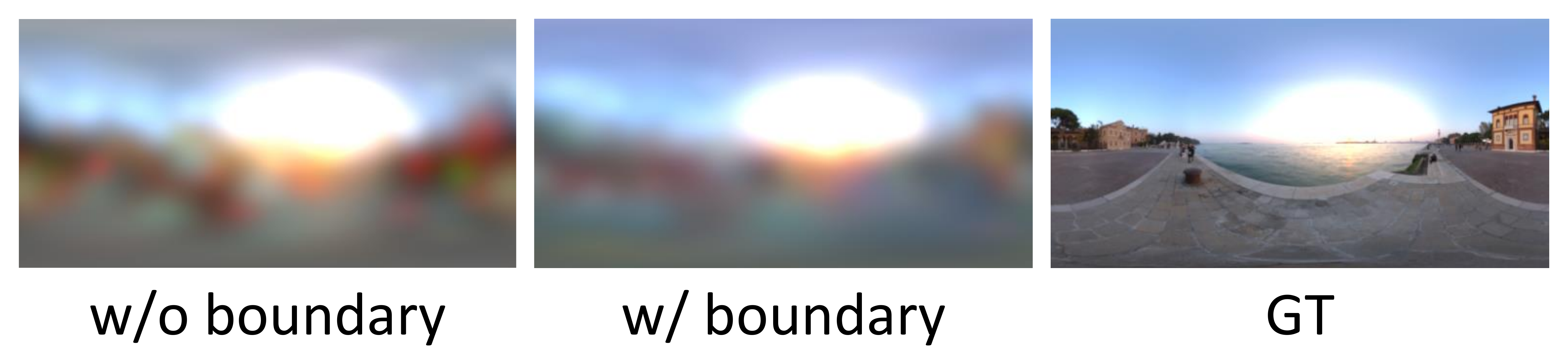}
  \caption{\textbf{Ablation on Boundary Lights}. Without boundary lights, optimized environment maps have noticeable artifacts.}
  \label{fig:ablation_boundary}
\end{figure}

\begin{figure*}[ht]
  \centering
  \includegraphics[width=1\linewidth]{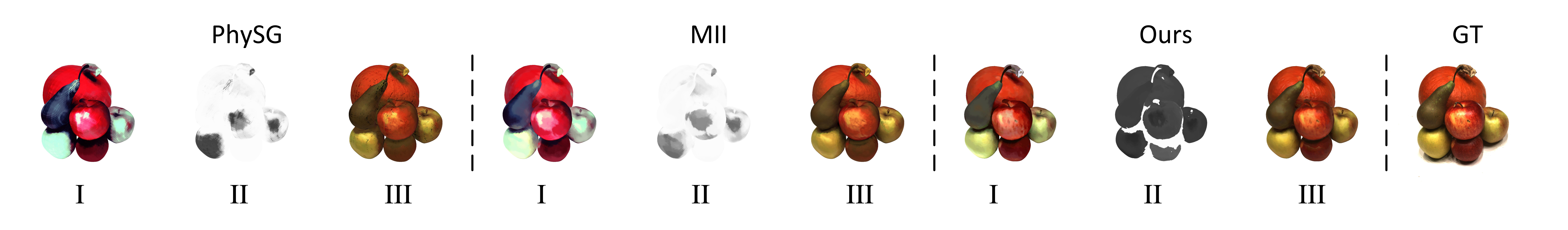}
  \vspace{-5mm}
  \caption{\textbf{Qualitative Comparisons on DTU}~\cite{c14}. Even for the real-world dataset, our method can still decompose reasonable albedo (\rom{1}), roughness(\rom{2}), and realistic rendering result (\rom{3}). In contrast, previous methods~\cite{c4,c10} struggle to preserve consistent roughness and albedo. Our method generalizes well to challenging real-world dataset, producing reasonable albedo (\rom{1}), roughness(\rom{2}), and realistic rendering result (\rom{3}). In contrast, the baselines~\cite{c4,c10} struggle to predict consistent roughness and albedo.}
  \label{fig:qualitative dtu}
\end{figure*}

\subsection{Ablation Studies}
\label{sec:ablation}
\noindent\textbf{Sphere Tracing Refinement}.
By allowing a large threshold to reduce tracing iterations, our sphere tracing refinement significantly reduces the time-consuming geometry initialization training time from 22 to 17 hours. More details can be found in the supplementary.


\noindent\textbf{Integral Weight}. 
As discussed in~\cref{sec:indirect light}, we estimate each indirect light by aggregating environment lights with integral weights predicted by an MLP. Below, we conduct an ablation study to show its advantage towards uniform weights. Specifically, We estimate indirect lights with learned and uniform weights, respectively, and show the rendered images in Fig.~\ref{fig:visual}a with their error maps (\ie differences compared to ground truth images) in Fig.~\ref{fig:visual}b. With learned weights, errors are reduced in the shading area (\eg the area around the hot dogs) compared to using uniform weights. Those shading areas are typically dominated by indirect light. Moreover, the last two rows of \cref{tab:novel view} quantitatively demonstrate the learned weight's superiority.


\noindent\textbf{Spherical Gaussian Number}. The number of environment lights determines how smoothly the lights are modeled. In the right of \cref{fig:material edit}, the quality of light estimation improves when Spherical Gaussians are used.
However, MSE only decrease from 0.0120
to 0.0116 when the number increases from 128 to 256.
Thus, we choose 128 environment lights as a trade-off between quality and memory efficiency.

\noindent\textbf{Boundary Lights}. The ablation model learned without boundary lights produces blurry environment maps (See~\cref{fig:ablation_boundary}) and significantly less accurate illumination prediction (See~\cref{tab:Light Estimation}). This shows that
boundary lights play a crucial role in accurate illumination estimation and verify the contribution of a differentiable boundary light learning module. 
Furthermore, the model learned with boundary lights shows slight but consistent improvement in albedo estimation across scenes (See ~\cref{tab:albedo}). However, we do not observe a noticeable advance in roughness prediction. More analysis on this term is in the supplementary.



\subsection{Evaluation on Real-World Images}
\label{sec:real}
We demonstrate the generalization ability of the proposed method by applying it to the DTU MVS dataset~\cite{c14}. 
We select objects with diverse materials for these experiments, including a bear toy, an owl statue, and a pile of fruits with heavy occlusion.
We estimate the camera pose by using the COLMAP method~\cite{c15}. 
Photos in the DTU dataset~\cite{c14} are taken in a black room with LED above the scenes. 
As a result, the environment illumination is not from an infinitely far distance as assumed in \cref{sec:sg} and is challenging for inverse rendering. 
We further compare our method with PhySG~\cite{c10} and MII~\cite{c4} on real-world data in \cref{fig:qualitative dtu}. Though there is no ground-truth material and roughness in the DTU~\cite{c14} as a reference, it is worth noting that our method obtains plausible roughness and albedo while the baselines can barely produce reasonable decomposition. Moreover, the details learned by our model, such as the patterns on the surface of the fruit, are more realistic than the baselines. 
More comparisons of the owl statue and the bear toy can be found in the supplementary.

\subsection{Discussions}
Self-supervised inverse rendering is a highly ill-posed problem. Though our method achieves competitive performance compared to SOTAs, it cannot model highly reflective objects (e.g., mirrors) without prior such as illumination~\cite{lyu2022neural,tiwary2022orca} or expensive Monte-Carlo sampling~\cite{,liu2023nero}. Our method also fails to model delicate surfaces such as hairs and furs due to the limited geometry representation capability of SDFs. We discuss more limitations and assumptions used for the physical rendering in the supplementary and leave them in future works.

\section{Conclusion}

In this paper, we introduce an inverse rendering framework that simultaneously predicts the illumination, geometry, and material of a scene from multi-view images. 
Through explicitly tracing and estimating indirect lights at each surface point, our approach effectively captures the reflection in the scene. 
We demonstrate the effectiveness of our method for inverse rendering, material editing, and free-viewpoint relighting on both real and synthetic datasets, achieving favorable performance against state-of-the-art inverse rendering models. 


{
    \small
    \bibliographystyle{ieeenat_fullname}
    \bibliography{main}
}

\end{document}